\documentclass[11pt,a4paper]{article}
\usepackage[hyperref]{emnlp2020}
\usepackage{times}
\usepackage{latexsym}
\usepackage{multirow}
\usepackage{multicol}
\usepackage{graphicx}
\usepackage{amsmath}
\usepackage{CJKutf8}
\usepackage[T1]{fontenc}

\usepackage{microtype}

\aclfinalcopy 
\def\aclpaperid{3299} 

\title{Improving the Efficiency of Grammatical Error Correction with \\ Erroneous Span Detection and Correction}

  \author{Mengyun Chen$^1$\thanks{\ \ This work was done during the author's internship at Microsoft Research Asia.}~\thanks{\ \ Co-first authors with equal contributions.}~~~~Tao Ge$^2$\footnotemark[2]~~~~Xingxing Zhang$^2$~~~~Furu Wei$^2$~~~~Ming Zhou$^2$\\
 $^1$ East China Normal University; $^2$ Microsoft Research Asia  \\
{\tt dreamcloud.chen@gmail.com}; \\ {\tt \{tage,xizhang,fuwei,mingzhou\}@microsoft.com}
}

\date{}

\begin{document}
\maketitle
\begin{abstract}
We propose a novel language-independent approach to improve the efficiency for Grammatical Error Correction (GEC) by dividing the task into two subtasks: Erroneous Span Detection (ESD) and Erroneous Span Correction (ESC). ESD identifies grammatically incorrect text spans with an efficient sequence tagging model. Then, ESC leverages a seq2seq model to take the sentence with annotated erroneous spans as input and only outputs the corrected text for these spans. Experiments show our approach performs comparably to conventional seq2seq approaches in both English and Chinese GEC benchmarks with less than 50\% time cost for inference.
\end{abstract}

\section{Introduction}\label{sec:intro}

Due to a growing number of error-corrected parallel sentences available in recent years, sequence-to-sequence (seq2seq) models with the encoder-decoder architecture \cite{bahdanau2014neural,sutskever2014sequence,luong2015effective} have become a popular solution to GEC, which take the source (original) sentence as input and output the target (corrected) sentence. Although auto-regressive seq2seq models facilitate correction for various grammatical errors and perform well, they are not efficient enough for GEC. As previous work \cite{zhao2019improving} points out, seq2seq models take most decoding steps to copy grammatically correct text spans from the source to the target during inference, which is the main efficiency bottleneck. If the time for the copying operations can be saved, the efficiency should be much improved.


\begin{figure*}[t]
\centering
\includegraphics[width=12cm]{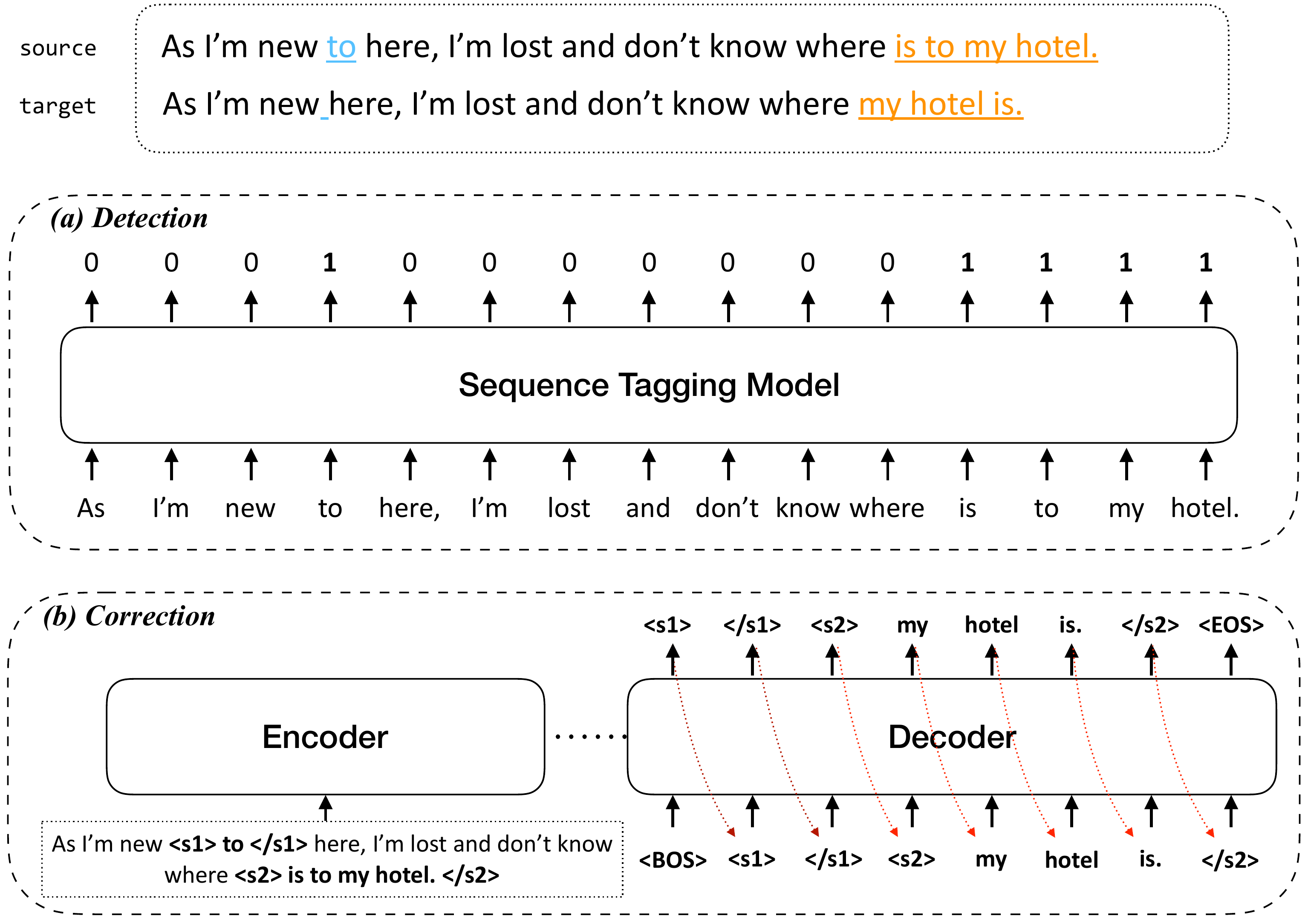}
\caption{An overview of erroneous span detection (ESD) and erroneous span correction (ESC). The detection model is a sequence tagging model, while the correction model is a seq2seq model but only outputs the corrected texts for annotated spans (i.e., $\langle s1 \rangle$ \textit{to} $\langle /s1 \rangle$ and $\langle s2 \rangle$ \textit{is to my hotel.} $\langle /s2 \rangle$). For example, the text span ``\textbf{is to my hotel.}'' is identified as incorrect and then edited into ``\textbf{my hotel is.}''.\label{fig:overview}}\vspace{-0.1cm}
\end{figure*}

With this motivation, we propose a simple yet novel language-independent approach to improve the efficiency of GEC by dividing the task into two subtasks: Erroneous Span Detection (ESD) and Erroneous Span Correction (ESC), shown in Figure \ref{fig:overview}. In ESD, we use an efficient sequence tagging model to identify the text spans that are grammatically incorrect in the source sentence, as Figure \ref{fig:overview}(a) shows. Then, we feed the sentence with erroneous span annotations to a seq2seq model for ESC. In contrast to conventional seq2seq approaches correcting the complete sentence, ESC only corrects the erroneous spans (see Figure \ref{fig:overview}(b)), which largely decreases the number of steps for decoding. Experiments in both English and Chinese GEC benchmarks demonstrate our approach performs comparably to the state-of-the-art transformer based seq2seq model with less than 50\% time cost for inference. Furthermore, our approach offers more flexibility to control correction, allowing us to adapt the precision-recall trade-off to various application scenarios.


\section{Related Work}
Recently, many approaches have been proposed to improve GEC performance. However, except those adding synthetic erroneous data~\cite{xie2018noising,ge2018fluency,grundkiewicz2019neural,kiyono2019empirical,zhou2019improving} and Wikipedia revision logs~\cite{lichtarge2019corpora} for training, most methods cause an increase in latency. For example, language model and right-to-left (R2L) rescoring~\cite{grundkiewicz2019neural,kiyono2019empirical} not only take time to rescore but also slow down the correction model with a larger beam size during inference; multi-round (iterative) decoding~\cite{ge2018fluency,ge2018reaching,lichtarge2019corpora} needs to repeatedly run the model; BERT-fuse~\cite{kaneko2020encoder} adds extra computation for model fusion.


In contrast to extensive studies on GEC performance, little work focuses on improving the efficiency of GEC models until the last years. One branch of the work is language-dependent approaches, like PIE \cite{awasthi2019parallel} and GECToR \cite{omelianchuk2020gector}. They predict a sequence of token-level edit operations including a number of manually designed language-specific operations like changing verb forms (e.g., \textit{VBZ}$\to$\textit{VBD}) and prepositions (e.g., \textit{in}$\to$\textit{on}). 
However, they are difficult to be adapted to other languages. The other branch is language-independent models like LaserTagger~\cite{malmi2019encode}. They learn a vocabulary of edit operations from training data and thus can work for any language. However, their performance is inferior to their seq2seq counterpart. Our approach combines the advantages of both branches, which is language-independent and performs comparably to the state-of-the-art seq2seq approach with efficient inference.

\begin{table*}[!t]
\centering
\small
\begin{tabular}{l|c|c|ccc|ccc}
\hline
\multirow{3}{*}{\textbf{Model}} & \multicolumn{1}{l|}{\multirow{3}{*}{\textbf{Pretrained}}} & \multicolumn{1}{l|}{\multirow{3}{*}{\textbf{Faster/Slower}}} & \multicolumn{3}{c|}{\multirow{2}{*}{\textbf{CoNLL-14 ($M^2$)}}} & \multicolumn{3}{c}{\multirow{2}{*}{\textbf{BEA-19 (ERRANT)}}} \\
                                & \multicolumn{1}{l|}{}                                     & \multicolumn{1}{l|}{}                                        & \multicolumn{3}{c|}{}                                   & \multicolumn{3}{c}{}                                 \\
                                & \multicolumn{1}{l|}{}                                     & \multicolumn{1}{l|}{}                                        & $P$       & $R$      & $F_{0.5}$    & $P$      & $R$      & $F_{0.5}$    \\ \hline
\it Seq2seq                         & No                                                        & -                                                            & 64.9             & 26.6            & \bf 50.4               & 57.3            & \bf  41.5            & 53.2              \\ 
\textit{Levenshtein Transformer}$^\star$~\cite{gu2019levenshtein} & No & Faster & 39.9 &  24.4 & 35.4 & 32.2 & 39.2 & 33.4 \\
\textit{Levenshtein Transformer}$^\star$~(distillation) & No & Faster & 53.1 &  23.6 & 42.5 & 45.5 & 37.0 & 43.5 \\
\textit{LaserTagger}$^\star$~\cite{malmi2019encode} & No & Faster & 50.9 & \bf  26.9 & 43.2 & 53.4 & 38.5 & 49.6 \\
Our approach                    & No                                                        & Faster                                                       & \bf 66.0             & 24.7            & 49.5               & \bf 62.7            & 38.6            & \bf 55.7              \\ \hline
\it Seq2seq                         & Yes                                                       & -                                                            & 69.4             & 42.5            & \bf 61.5               & 66.7            & \bf 61.3            & 65.5              \\ 
\textit{PRETLarge}~\cite{kiyono2019empirical}                       & Yes                                                       & -                                                            & 67.9             & \bf 44.1            & 61.3               & 65.5            & 59.4            & 64.2              \\ 
\textit{PIE}~\cite{awasthi2019parallel}                             & Yes                                                       & Faster                                                       & 66.1             & 43.0            & 59.7               & \underline{58.0}               & \underline{53.1}   &  \underline{56.9}                 \\ 
Our approach                    & Yes                                                       & Faster                                                       & \bf 72.6             & 37.2            & 61.0               & \bf 70.4            & 55.9            & \bf 66.9              \\ \hline
\textit{BERT-fuse GED}~\cite{kaneko2020encoder}                   & Yes                                                       & Slower                                                       & 69.2             & 45.6            & 62.6               & 67.1            & 60.1            & 65.6              \\ 
\textit{BERT-fuse GED+R2L}~\cite{kaneko2020encoder}               & Yes                                                       & Slower                                                       & \bf 72.6             & \bf 46.4            & \bf 65.2               & \bf  72.3            & 61.4            & \bf 69.8              \\ 
\textit{PRETLarge+SSE+R2L}~\cite{kiyono2019empirical}               & Yes                                                       & Slower                                                       & 72.4             & 46.1            & 65.0               & 72.1            & \bf 61.8            & \bf 69.8              \\ 
\textit{UEDIN-MS}~\cite{grundkiewicz2019neural}                        & Yes                                                       & Slower                                                       & -                & -               & 64.2               & \bf 72.3            & 60.1            & 69.5              \\ \hline
\end{tabular}
\caption{Performance in English GEC benchmarks (i.e., CoNLL-14 and BEA-19 test). \textit{Seq2seq} is our implemented seq2seq model based on \textbf{Transformer (big)} architecture, which is also the baseline for speed comparison (i.e., \textbf{Faster/Slower} in the table). The column \textbf{Pretrained} indicates whether the model is pretrained with synthetic or additional (e.g., Wikipedia revision logs) error-corrected data. $^\star$ indicates the models are implemented by us with the released codes of the original papers, trained and evaluated on the BEA-19 setting. The underlines indicate the scores are evaluated by us for the released model on the BEA-19 test data.  
\label{tab:base_result}}
\end{table*}

\section{Erroneous Span Detection}\label{sec:detection}\vspace{-0.1cm}
To identify incorrect spans, we use a binary sequence tagging model in which tag \textbf{0} means the token is in a correct span; while tag \textbf{1} means the token is in a grammatically incorrect span that needs to be edited, as shown in Figure \ref{fig:overview}(a). In order to train the tagging model, we align\footnote{Alignment can be solved by dynamic programming like Levenshtein distance. We here use ERRANT (\url{https://github.com/chrisjbryant/errant}) for alignment.} tokens across the source and target sentence in training data. With token alignment, we can identify the text spans that are edited and thus can annotate the edited text spans in the original sentences as erroneous spans.



\section{Erroneous Span Correction}\label{sec:correction}\vspace{-0.1cm}
With ESD, we can identify grammatically incorrect text spans in a sentence. If a sentence is identified as error-free, we take no further action; otherwise, we annotate the incorrect spans and use the ESC model to correct them, shown in Figure \ref{fig:overview}(b).



To avoid ESC being misled by span detection errors from ESD during inference, we randomly select text spans in the similar way to SpanBERT~\cite{joshi2019spanbert} instead of only using gold erroneous spans in training data, to train the ESC model. In this way, the ESC model will see a large variety of span annotations and learn how to correct during training, and thus its robustness is improved: even if the detected spans during inference are not exactly accurate, the ESC model will not easily fail. With token alignment across the source and target sentence in GEC training data, we can generate training instances with span annotations and corrections like the example in Figure \ref{fig:overview}(b) for ESC.

\begin{table*}[t]
\centering
\small
\scalebox{0.97}{
\begin{tabular}{l|cccc|ccc|cccc|ccc}
\hline
\multirow{3}{*}{\textbf{Model}} & \multicolumn{7}{c|}{\textbf{CoNLL-14 (1,312)}}                                                   & \multicolumn{7}{c}{\textbf{NLPCC-18 (2,000)}}                                                   \\ \cline{2-15} 
                                & \multicolumn{4}{c|}{\textbf{Time (in second)}}                  & \multicolumn{3}{c|}{\textbf{Performance}} & \multicolumn{4}{c|}{\textbf{Time (in second)}}                  & \multicolumn{3}{c}{\textbf{Performance}} \\ 
                                & \textbf{1} & \textbf{8} & \textbf{16} & \textbf{32} & $P$   & $R$   & $F_{0.5}$  & \textbf{1} & \textbf{8} & \textbf{16} & \textbf{32} & $P$   & $R$   & $F_{0.5}$  \\ \hline
\textit{Seq2seq}                         & 363        & 85         & 51          & 33          & 64.9         & \bf 26.6         & \bf 50.4        & 690        & 166        & 101         & 63          & 36.9         & 14.4         & 28.1        \\ 
\textit{Levenshtein Transformer}         & \bf 125        & \bf 24         & \bf 19          & \bf 14          & 53.1        & 23.6         & 42.5        & \bf 224        &  64         & 41          & 31          & 24.9       & \bf 15.0         & 22.0       \\ 
Our approach                    &  137        & 34         &  21          &  16          & \bf 66.0         & 24.7         & 49.5        &  253        & \bf 60         & \bf 39          & \bf 29          & \bf 37.3         &  14.5         & \bf 28.4        \\  \hline
\textit{Seq2seq (tensor2tensor)}         & 680        & 138        & 97          & 85          & 58.7         & 30.5         & 49.5        & 1292       & 227        & 141         & 92          & 41.0         & 10.8         & 26.3        \\ 
\textit{PIE}        & 66         & 52         & 51          & 48          & -         & -       & -        & -          & -          & -           & -           & -            & -            & -           \\ 
\textit{LaserTagger}                     & \bf 23         & \bf 12         & \bf 9           & \bf 8           & 50.9         & 26.9         & 43.2        & \bf 34         & \bf 16         & \bf 14          & \bf 13          & 25.6         & 10.5         & 19.9        \\ \hline
\end{tabular}
}
\caption{Performance and total inference time of models without pretraining under various batch sizes (1/8/16/32) using 1 Nvidia V100 GPU with CUDA 10.2 in the English (CoNLL-14: 1,312 sentences) and Chinese (NLPCC-18: 2,000 sentences) GEC test sets. The top group of models is implemented with Pytorch, while the bottom group is implemented with Tensorflow, thus their inference time cannot be compared. The performance of PIE in CoNLL-14 is not reported because it is pretrained with synthetic data and thus unfair to be compared here. Also, PIE has no result in NLPCC-18 because it is specific for English and difficult to be generalized to other languages.\label{tab:time}}\vspace{-0.25cm}
\end{table*}

\begin{table*}[t]
\centering
\small
\scalebox{1}
{
\begin{tabular}{l|c|ccc|ccc}
\hline
\multirow{2}{*}{\textbf{Model}}                & \multirow{2}{*}{\textbf{\#Sent for ESC/seq2seq}} & \multicolumn{3}{c|}{\textbf{Batch size}} & \multirow{2}{*}{$P$} & \multirow{2}{*}{$R$} & \multirow{2}{*}{$F_{0.5}$} \\ 
                   &                            & 1           & 8           & 16          &                             &                             &                             \\ \hline
ESD (base) + ESC & 872 &      \bf  137 (23+114)      &   \bf   34 (5+29)        &   \bf   21 (3+18)        &          66.0             &               24.7 &             49.5                \\
ESD (base) + seq2seq & 872   &    284 (23+261)         &     66 (5+61)        &       41 (3+38)      &          \bf 67.0                   &  \bf 25.1      &  \bf   50.2                         \\ \hline
ESD (large) + ESC &   935 & \bf   167 (43+124)    &    \bf 40 (8+32)       &   \bf   25 (6+19)        &        \bf       67.2               &  \bf          26.4        & \bf  51.3       \\
ESD (large) + seq2seq  & 935  &     318 (43+275)        &      77 (8+69)       &            47 (6+41)    &                 66.6            &          25.5                   &        50.3                     \\ \hline
seq2seq & 1,312     &      363       &      85       &      51       &        64.9                     &           26.6                  &                50.4            \\ \hline
\end{tabular}
}\vspace{-0.1cm}
\caption{In-depth time cost (in second) analysis in CoNLL-14 which contains 1,312 test sentences. (base) and (large) indicate that the ESD models are fine-tuned from the Roberta base and large models respectively. ESD + seq2seq is implemented as follows: ESD first identifies the sentences that have grammatical errors, then the seq2seq baseline model only corrects these sentences. The column \textbf{\#Sent for ESC/seq2seq} shows the actual number of sentences ESC/seq2seq processed. For the time cost in the brackets such as (23+114), the first term (e.g., 23) is the time cost by the ESD model while the last term is the cost (e.g., 114) by the other parts.\label{tab:depth_time}}\vspace{-0.2cm}
\end{table*}

\section{Experiments}\vspace{-0.15cm}
\subsection{Experimental Setting}\vspace{-0.05cm}
Following recent work in English GEC, we conduct experiments in the same setting with the restricted track of the BEA-2019 GEC shared task~\cite{bryant2019bea}, using FCE~\cite{yannakoudakis2011new}, Lang-8 Corpus of Learner English~\cite{mizumoto2011mining}, NUCLE~\cite{dahlmeier2013building} and W\&I+LOCNESS~\cite{granger1998computer,bryant2019bea} as training data. We use CoNLL-2013 test set as the dev set to choose the best-performing models, and evaluate on the well-known GEC benchmark datasets: CoNLL-2014~\cite{ng2014conll} and BEA-2019 test set with the official evaluation scripts (m2scorer\footnote{\url{https://github.com/nusnlp/m2scorer}} for CoNLL-14, ERRANT for BEA-19). As previous work~\cite{grundkiewicz2019neural} trained with synthetic data, we synthesize 260M sentence pairs in the same way to try pretraining ESD and ESC. Also, we verify in Chinese GEC whether our approach can be adapted to other languages. We follow the setting of NLPCC-2018 Chinese GEC shared task \cite{zhao2018overview}, using its official training\footnote{We sample 5,000 training instances as the dev set.} and evaluation datasets.

We fine-tune Roberta \cite{DBLP:journals/corr/abs-1907-11692} base and Chinese Bert \cite{Bert} base model for English and Chinese ESD respectively. For ESC, we train a Transformer (big) model \cite{vaswani2017attention}, using an encoder-decoder shared vocabulary of 32K Byte Pair Encoding~\cite{sennrich2015neural} tokens for English and 8.4K Chinese character for Chinese. During inference, ESC decodes with a beam size of 5. We include more details of models, training and inference in the Appendix. 

\subsection{Experimental Results}
Table \ref{tab:base_result} shows the performance of our approach and recent models in English GEC. Although the non auto-regressive models like Levenshtein Transformer and LaserTagger are faster than the seq2seq baseline, their performance is not desirable. Among the models without pretraining (\textbf{top group}), our approach is the only one that performs comparably to the \textit{Seq2seq} baseline with faster inference. When we add the synthetic data to pretrain the ESD and ESC models, our approach's results are much improved, yielding state-of-the-art results among the models with good inference speed (\textbf{middle group}). Though our approach underperforms the best systems (\textbf{bottom group}) which improve results through various methods (e.g., model fusion, ensemble decoding and rescoring) that seriously hurt efficiency, it is much more efficient and thus applicable in real world applications.

\begin{figure}[t]
    \centering
    \includegraphics[width=8.2cm]{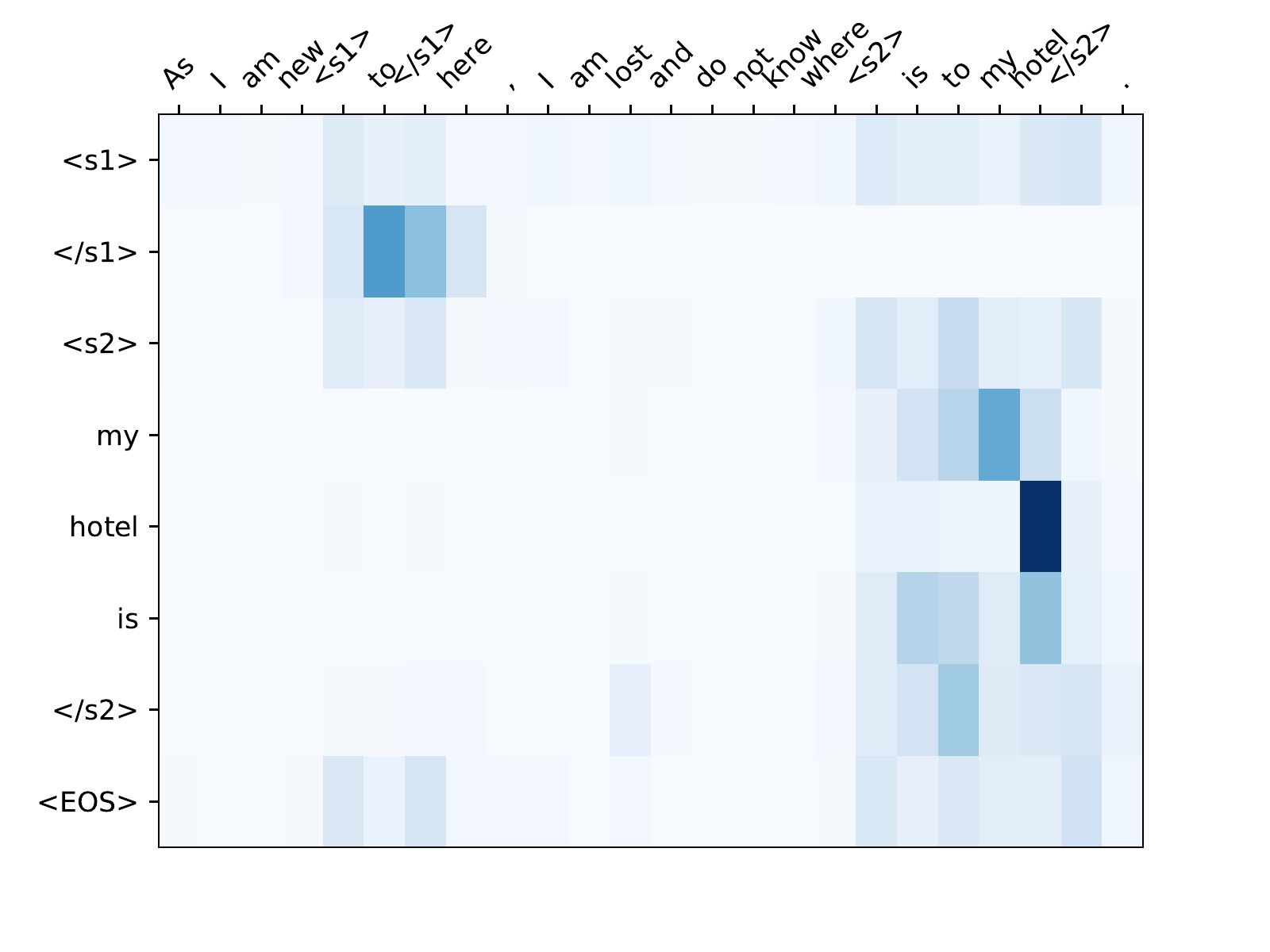}\vspace{-0.7cm}
    \caption{Attention Heatmap of ESC.}\vspace{-0.5cm}
    \label{fig:heatmap}
\end{figure}



Table \ref{tab:time} compares the inference time of various approaches. Compared to the \textit{Seq2seq} implementation in Pytorch-fairseq, our approach saves over 50\% time cost. It is notable that among the implementations in Table \ref{tab:time}, LaserTagger is the most efficient though its results are not good enough. Also, our approach consistently achieves comparable performance with the \textit{Seq2seq} baseline in Chinese GEC, demonstrating that our approach can be easily adapted to other languages. 


We further analyze the corresponding time cost of ESD and ESC. Table \ref{tab:depth_time} shows that ESD is much faster than the auto-regressive ESC model. It not only efficiently filters out error-free sentences to save effort for the following process, but also pinpoints incorrect spans, allowing ESC to only focus on correcting the spans and reduce the decoding steps, as shown in Figure \ref{fig:heatmap}. For the 872 sentences ESD (base) identifies as incorrect, the total number of decoding time steps (in the best beam) of ESC is 7,647, accounting for the efficiency improvement over the seq2seq model whose corresponding decoding steps are 21,065. Furthermore, if we use a larger ESD model (Roberta large), we observe better results with a still marked reduction in time cost compared to the baseline. More detailed qualitative studies and analyses of ESD are presented in the supplementary notes due to the space limitation.

\begin{table}[t]
\centering
\small
\begin{tabular}{c|c|c|c|c|c|c}
\hline
threshold & 0.2 & 0.3 &0.4 & 0.5 & 0.6 & 0.7 \\ \hline
$P$ & 62.4 & 63.8 & 64.8 & 66.0 & 66.2 & \bf 67.0 \\ 
$R$ & \bf 27.4 & 26.8 & 25.6 & 24.7 & 23.4 & 21.7 \\
$F_{0.5}$ & 49.7 & \bf 50.0 & 49.6 & 49.5 & 48.4 & 47.2 \\
\hline
\end{tabular}
\caption{As the probability threshold of ESD increases, precision increases while recall drops in CoNLL-14.\label{tab:threshold}}
\end{table}

Besides its efficiency advantage, our approach offers more flexibility to control correction behavior during inference, making it adaptive to various real-world application scenarios. As shown in Table \ref{tab:threshold}, if the model is intended for high precision, we can increase the probability threshold for ESD so that it identifies the incorrect spans only when it is very confident; on the other hand, if we want the model to be aggressive for higher recall, we can simply decrease the threshold.

\section{Conclusion and Future Work}\vspace{-0.05cm}
We propose a novel language-independent approach to improve the efficiency of GEC. Our approach performs comparably to the state-of-the-art seq2seq model with a considerable reduction in inference time, and can be easily adapted to other languages and offer more flexibility to control correction behavior (e.g., trading precision for recall).


Through our experiments in GEC, we verify the feasibility of span-specific decoding, which has been explored for text infilling~\cite{raffel2019exploring} and text rewriting. It is inspiring and promising to be generalized to more rewriting tasks, which will be studied as our future work.

\section*{Acknowledgments}
We thank Xin Sun and Xiangxin Zhou for their help with experiments. We thank the anonymous reviewers for their valuable comments. The corresponding author of this paper is Tao Ge.

\bibliographystyle{acl_natbib}
\bibliography{emnlp2020}

\appendix

\section{Experiment Details}
Table \ref{tab:datasets} describes the details of datasets used for English GEC. Except the sythetic data, all the data can be found at the website\footnote{\url{https://www.cl.cam.ac.uk/research/nl/bea2019st/}} of the BEA-19 shared task. The synthetic data is generated from English Wikipedia\footnote{\url{https://en.wikipedia.org/}}, English Gigaword~\cite{parker2011english} and Newscrawl\footnote{\url{http://data.statmt.org/news-crawl/en/}} as the previous work~\cite{ge2018fluency,zhang2019sequence,kiyono2019empirical,grundkiewicz2019neural} did, using back translation and sentence corruption. Specifically, we train a transformer (base) model~\cite{vaswani2017attention} for back translation using the training data of the restricted track in the BEA-19 shared task. For sentence corruption, we follow \newcite{edunov2018understanding} to randomly insert, delete, replace and swap adjacent tokens in a sentence.

Hyper-parameters for the ESD and ESC model for English GEC are listed in table \ref{tab:param_esd} and table \ref{tab:param_esc}. The hyper-parameters for Chinese GEC are almost the same except that the ESD model is fine-tuned from Chinese Bert \cite{Bert} base model.

At last, we highlight that the Levenshtein Transformer baselines in this paper are implemented using the master branch of fairseq (previous versions may have different reproduced results).

\begin{table}[t]
\centering
\small
\begin{tabular} {llll} 
\hline
\bf Corpus  & Sent. & Tok.  & Usage \\
\hline
\bf Synthetic data &  256M     & 4.88B      &  Pretraining\\
\hline
\bf FCE     & 32.0K & 0.5M  & Fine-tuning\\
\bf Lang-8  & 1.04M & 13.0M & Fine-tuning\\
\bf NUCLE   & 57.1K & 1.2M  & Fine-tuning\\
\bf W\&I+LOCNESS    & 34.3K & 0.6M  & Fine-tuning\\
\hline
\bf CoNLL-13& 1,381 & 29.2K & Development \\
\hline
\bf BEA-19  & 4,477 & 89.3K  & Test \\    
\bf CoNLL-14& 1,312 & 30.1K  & Test \\
\hline
\end{tabular}
\caption{Statistics of the datasets used for pretraining, fine-tuning and evaluation.\label{tab:datasets}}
\end{table}

\begin{table}[t]
\centering
\small
\begin{tabular} {lr} 
\hline
Configurations	         &	Values				\\
\hline
\multicolumn{2}{c}{\bf Pretraining} \\
\hline
Model Architecture 	    & Roberta (base)    \\
                        & ~\cite{DBLP:journals/corr/abs-1907-11692} \\
Number of parameters    & 125M \\
Number of epochs		& 5				\\
Devices                 & 8 Nvidia V100 GPU      \\
Max tokens per GPU		& 12000					\\
Optimizer 				& Adam 	\\
						& ($\beta_1$=0.9, $\beta_2$=0.98, $\epsilon$=$1\times10^{-6}$)	\\
                        & ~\cite{Adam} \\
Learning rate 			& $3\times10^{-5}$ \\
Learning rate scheduler & inverse sqrt \\ 
warmup                  & 8000 \\
weight decay            & 0.1 \\
Dropout 				& 0.3 \\
\hline
\multicolumn{2}{c}{\bf Fine-tuning} \\
\hline
Number of epochs		& 50				\\
Devices                 & 4 Nvidia V100 GPU      \\
Max tokens per GPU		& 8000					\\
Learning rate 			& $1\times10^{-5}$ \\
warmup                  & 4000 \\
Dropout 				& 0.2 \\
\hline
\end{tabular}
\caption{Hyper-parameters values of ESD during the pretraining and fine-tuning. \label{tab:param_esd}} 
\end{table}

\begin{table}[t]
\centering
\small
\begin{tabular} {lr} 
\hline
Configurations	         &	Values				\\
\hline
\multicolumn{2}{c}{\bf Pretraining} \\
\hline
Model Architecture 	    & Transformer (big) \\
                        & ~\cite{vaswani2017attention}\\
Number of parameters    & 209M \\
Number of epochs 		& 5					\\
Devices                 & 8 Nvidia V100 GPU      \\
Max tokens per GPU		& 12000					\\
Optimizer 				& Adam 					\\
						& ($\beta_1$=0.9, $\beta_2$=0.98, $\epsilon$=$1\times10^{-8}$)	\\
                        & ~\cite{Adam} \\
Learning rate 			& $5\times10^{-4}$			\\
Learning rate scheduler & polynomial decay \\
Warmup                  & 8000 \\
weight decay            & 0.0 \\
Loss Function 			& label smoothed cross entropy \\
						& (label-smoothing=0.1) \\
						& ~\cite{label_smooth} \\
Dropout					& 0.3	\\
\hline
\multicolumn{2}{c}{\bf Fine-tuning} \\
\hline
Number of epochs 		& 30					\\
Devices                 & 4 Nvidia V100 GPU      \\
Max tokens per GPU		& 5120					\\
Learning rate 			& $3\times10^{-4}$		\\
Warmup                  & 4000 \\
Beam search 			& 5 					\\
\hline
\end{tabular}
\caption{Hyper-parameters values of ESC during the pretraining and fine-tuning.\label{tab:param_esc}} 
\end{table}

\begin{table}[t]
\centering
\small
\scalebox{0.85}{
\begin{tabular}{l|ccc|ccc}
\hline
\multirow{3}{*}{\textbf{ESD Model}} & \multicolumn{3}{c|}{\multirow{2}{*}{\textbf{Annotation 1}}} & \multicolumn{3}{c}{\multirow{2}{*}{\textbf{Annotation 2}}} \\
 & \multicolumn{3}{c|}{}    & \multicolumn{3}{c}{}  \\
 & P  & R   & $F_{0.5}$  & P  & R  &$F_{0.5}$ \\ 
 \hline
 ESD (base) & \textbf{52.4} & 35.3 & 47.8 & 63.4 & 31.3 & 52.6 \\
 ESD (large) & 49.5 & 40.0 & 47.2 & 61.4 & 36.4 & 54.0 \\
 ESD (base+pretrained) & 50.9 & \textbf{40.1} & \textbf{48.3} & \textbf{63.7} & \textbf{36.8} & \textbf{55.6} \\
 \hline
\end{tabular}
}
\caption{The performance of ESD on the two official annotations for the CoNLL-14 shared task test dataset. 
 \label{tab:ESD_performance}}
\end{table}


\section{ESD performance}
 We followed the previous work ~\cite{rei-2017-semi,rei-yannakoudakis-2017-auxiliary,kaneko-etal-2017-grammatical} in Grammatical Error Detection (GED), using token-level \textit{precision}, \textit{recall} and $F_{0.5}$ to evaluate  our ESD model. Table \ref{tab:ESD_performance} shows the results in CoNLL-14.

\section{Examples}
In table \ref{tab:examples} and table \ref{tab:zh_examples}, we show examples that are corrected by our approach to demonstrate the effectiveness of our approach in practice. According to the results in these tables, it is clear that our approach can yield satisfying corrections without hurting fluency, which is consistent with our evaluation results in the JFLEG~\cite{napoles2017jfleg} test set with respect to GLEU\footnote{Our approach (without pretraining) achieves 53.0 GLEU, comparable to 52.7 by its seq2seq counterpart in JFLEG.}~\cite{napoles2015ground} -- an automatic
fluency metric for GEC.

\begin{table*}[htbp]
\centering
\small
\scalebox{1}
{	
\begin{tabular} {ll} 
\hline
\hline
\textbf{Source Sentence} & Instead, we will post a seed and tag our friends to inform this kind of changments.  \\
\textbf{Annotation}     & Instead, we will post a seed and tag our friends to \textbf{<s1> inform this kind of changments. </s1>}\\
\textbf{Correction}     & \bf <s1> inform them of these kinds of changes. </s1> \\
\textbf{Final Output}   & Instead, we will post a seed and tag our friends to inform them of these kinds of changes. \\
\hline
\hline

\textbf{Source Sentence} & Personally I feel that we still should take our responsibility to tell them the situation.  \\
\textbf{Annotation}     & \textbf{<s1> Personally I </s1>} feel that \textbf{<s2> we still should </s2>} take our responsibility to tell them the situation.\\
\textbf{Correction}     & \bf <s1> Personally , I </s1> <s2> we should still </s2> \\
\textbf{Final Output}   & Personally, I feel that we should still take our responsibility to tell them the situation.\\
\hline
\hline
\textbf{Source Sentence}& The law 's spirit also include the fairness.   \\
\textbf{Annotation}     & The law 's spirit \textbf{<s1> also include the fairness. </s1>}\\
\textbf{Correction}     &\bf  <s1> also includes fairness. </s1> \\
\textbf{Final Output}   & The law 's spirit also includes fairness. \\
\hline
\hline
\textbf{Source Sentence}& Above all, life is more important than secret. \\
\textbf{Annotation}     & Above all, life is more important \textbf{<s1> than secret. </s1>} \\
\textbf{Correction}     &\bf <s1> than a secret. </s1> \\
\textbf{Final Output}   & Above all, life is more important than a secret. \\
\hline
\hline
\textbf{Source Sentence}& So, they have to also prepare mentally. \\
\textbf{Annotation}     & So, \textbf{<s1> they have to also prepare </s1>} mentally. \\
\textbf{Correction}     &\bf <s1> they also have to prepare </s1> \\
\textbf{Final Output}   &  So, they also have to prepare mentally.\\
\hline
\hline
\textbf{Source Sentence}& To prevent the bigger problem to happen, it takes a lot of effort to take care of your body. \\
\textbf{Annotation}     & To prevent the bigger \textbf{<s1> problem to happen, it </s1>} takes a lot of effort to take care of your body. \\
\textbf{Correction}     &\bf <s1> problem from happening, it </s1> \\
\textbf{Final Output}   &  To prevent the bigger problem from happening, it takes a lot of effort to take care of your body.\\
\hline
\hline
\end{tabular}
}
\caption{Examples of our ESD \& ESC approach in English for GEC.   ESD first detects the grammatical incorrect text spans in the source sentence. Then the sentence with the erroneous span annotations (the \textbf{Annotation} row) are fed into the ESC model to generate the corresponding corrections (the \textbf{Correction} row) for the annotated spans. Finally, we replace the erroneous spans with the corresponding corrected text in ESC's outputs (the \textbf{Final Output} row).\label{tab:examples}}
\end{table*}

\begin{CJK*}{UTF8}{gbsn}

\begin{table*}[!htbp]
\centering
\small
\scalebox{1}
{
\begin{tabular} {ll} 
\hline
\hline
\textbf{Source Sentence}& 北京的空气太污染了，泛在北京的人一定要注意，别抽烟。 \\
\textbf{Annotation}     & 北京的空气\textbf{<s1>太污染了，泛在北</s1>} 京的人一定要注意，别抽烟。\\
\textbf{Correction}     &\bf <s1> 污染太严重了，在北</s1> \\
\textbf{Final Output}   &  北京的空气污染太严重了，在北京的人一定要注意，别抽烟。\\
\hline
\hline
\textbf{Source Sentence}& 因为几乎的人们还没感到污染对自己的直接的影响。 \\
\textbf{Annotation}     & 因\textbf{<s1>为几乎的人们还</s1>}没感到污染对自己的直接的影响。 \\
\textbf{Correction}     & \bf 	<s1>为人们几乎</s1> \\
\textbf{Final Output}   &  因为为人们几乎还没感到污染对自己的直接的影响。\\
\hline
\hline
\textbf{Source Sentence}& 列车、汽车，飞机等人类科技发展的结果也重大问题。 \\
\textbf{Annotation}     & 列车、汽车，飞机等人类科技发展的结果\textbf{<s1>也重大</s1>}问题。 \\
\textbf{Correction}     & \bf <s1>也存在重大</s1> \\
\textbf{Final Output}   &  列车、汽车，飞机等人类科技发展的结果也存在重大问题\\
\hline
\hline
\textbf{Source Sentence}& 中国，悠久的历史，灿烂的文化，真是在历史上最难忘的国家。 \\
\textbf{Annotation}     & 中国，悠久的历史，灿烂的文化，真\textbf{<s1>是在历</s1>}史上最难忘的国家。 \\
\textbf{Correction}     & \bf <s1>是历</s1> \\
\textbf{Final Output}   & 中国，悠久的历史，灿烂的文化，真是历史上最难忘的国家。 \\
\hline
\hline
\textbf{Source Sentence}& 以找到这些稳定的工作，我们有读书的必要 \\
\textbf{Annotation}     & \textbf{<s1>以找</s1>}到这些稳定的工作，我们有读书的必要。 \\
\textbf{Correction}     & \bf <s1>为了找</s1> \\
\textbf{Final Output}   &  为了找到这些稳定的工作，我们有读书的必要\\
\hline
\hline
\end{tabular}
}
\caption{Examples of our ESD \& ESC approach in Chinese for GEC.  \label{tab:zh_examples}} 
\end{table*}

\end{CJK*}

\end{document}